\documentclass[sigconf]{acmart}

\usepackage{multirow}

\renewcommand\footnotetextcopyrightpermission[1]{}
\AtBeginDocument{%
  }

\acmConference[MM '26] {Proceedings of the 34th ACM International Conference on Multimedia}{November 10--November 14, 2026}{Rio de Janeiro, Brazil.}

\acmSubmissionID{4926}

\begin{document}

\title{Personalized Image Aesthetic Assessment via Preference-rich Sample Mining and Cohort Merging}


\author{Zhichao Yang}
\authornote{Equal Contribution.}
\affiliation{%
	\institution{Xidian University}
	\city{Xi'an}
	\country{China}}
\email{yangzhichao@stu.xidian.edu.cn}

\author{Tianjiao Gu}
\authornotemark[1]
\affiliation{
  \institution{Xidian University}
  \city{Xian}
  \country{China}
}
\email{gutianjiao@stu.xidian.edu.cn}

\author{Zhixianhe Zhang}
\affiliation{
  \institution{Xidian University}
  \city{Xian}
  \country{China}
}
\email{zhixianhe.zhang@stu.xidian.edu.cn}

\author{Xiangfei Sheng}
\affiliation{
  \institution{Xidian University}
  \city{Xian}
  \country{China}
}
\email{xiangfeisheng@gmail.com}

\author{Pengfei Chen}
\affiliation{%
	\institution{Xidian University}
	\city{Xi'an}
	\country{China}}
\email{chenpengfei@xidian.edu.cn}

\author{Leida Li}
\authornote{Corresponding Author.}
\affiliation{%
	\institution{Xidian University}
	\city{Xi'an}
	\country{China}}
\email{ldli@xidian.edu.cn}


\begin{abstract}
  Personalized Image Aesthetic Assessment (PIAA) aims to predict aesthetic ratings of images that vary across individuals. The aesthetic preferences manifest to different extents across distinct visual stimuli and exhibit cohort-specific patterns. Motivated by the above fact, this paper presents a Multimodal Large Language Model (MLLM)-based approach, which models individual aesthetic preferences by \textbf{P}reference-\textbf{R}ich sample mining and \textbf{A}esthetically-resonant \textbf{C}ohort merging (\textbf{PRAC}). Specifically, PRAC first identifies preference-rich samples by analyzing both Collective Controversy and Personalized Deviation of images, maximizing the utility of limited user data. Based upon the preference-rich samples, cross-user preference similarities are measured by comparing preference embeddings. Then, a cohort-based model merging strategy, is proposed by aggregating preference patterns from aesthetically-resonant users, which further enhances the personalization for the target individual. Extensive experiments and comparisons on four benchmark PIAA databases demonstrate the superiority of the proposed PRAC model over the state-of-the-arts. The code and model will be public at \href{https://github.com/yzc-ippl/PRAC}{https://github.com/yzc-ippl/PRAC}.
\end{abstract}


\begin{teaserfigure}
  \centering
  \includegraphics[width=0.95\textwidth]{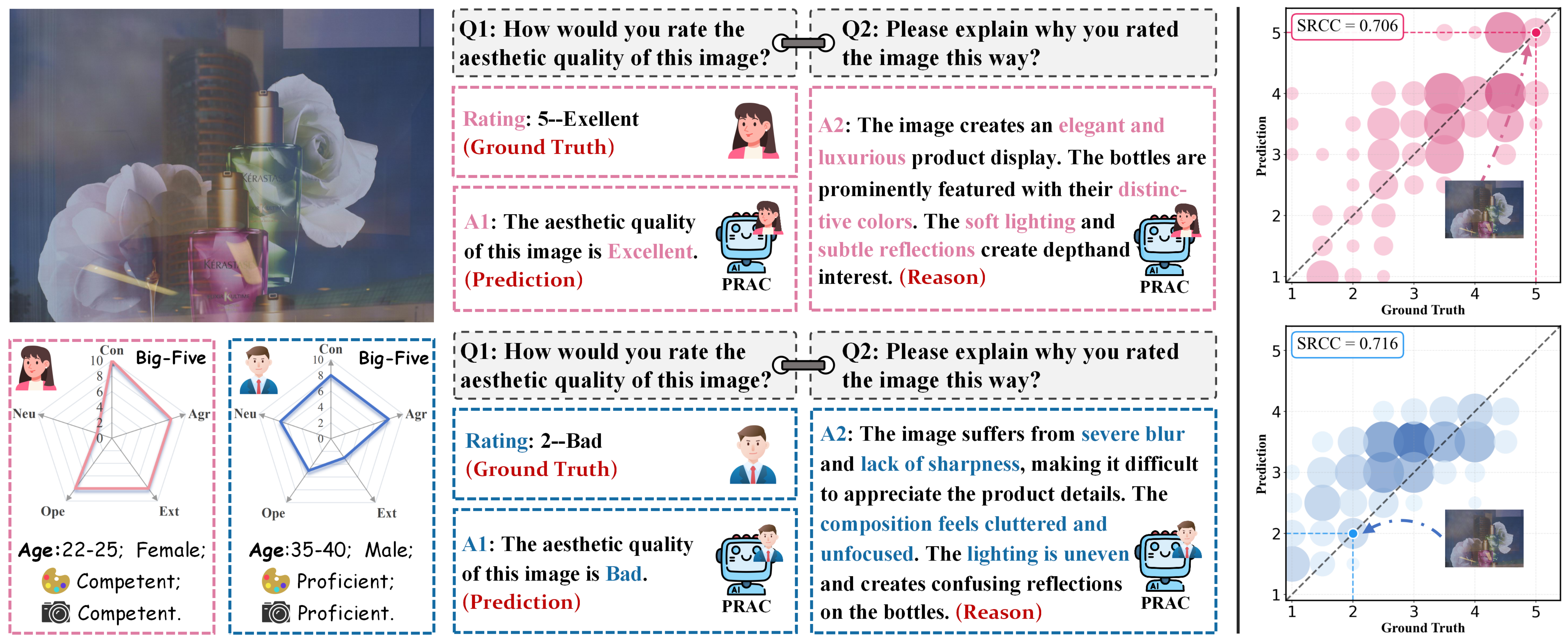}
  \caption{\textit{Beauty is in the eye of the beholder.} Individuals with distinct profiles demonstrate significant variations in aesthetic judgments of identical images. We propose PRAC, a Multimodal Large Language Model (MLLM)-based framework that accurately predicts aesthetic ratings aligned with individual preferences and offers interpretable rationales explaining divergent judgments.}
  \label{fig:teaser}
\end{teaserfigure}


\maketitle

\section{Introduction}
The widespread use of smartphones and social networks have dramatically transformed the consumption patterns of visual media. This shift has intensified user attention on image aesthetics, establishing image aesthetic assessment (IAA) as a prominent research topic \cite{deng2017image,yang2026fine,li2024towards}. Early studies primarily focused on generic IAA, which predicts average aesthetic ratings that reflect general judgments \cite{murray2012ava,yang2024semantics,huang2024aesbench,sheng2026fine}. However, \textbf{aesthetics is inherently subjective\textemdash different users may have substantially different aesthetic evaluations of the same image} (as illustrated in Figure \ref{fig:teaser}). This has driven growing interest in personalized image aesthetic assessment (PIAA), which models individual preferences by adapting to user-centric aesthetic tastes \cite{ren2017personalized,chen2025role,maerten2025lapis}. With the surging demand for customization, PIAA offers expanding applications including personalized image recommendations \cite{chen2024tailored,liu2024multimodal,sheng2025instructcrop}, user-centric photo enhancement \cite{zhang2022aesthetic,ni2022composition,sheng2026tuningiqa}, and personalized image generation \cite{xu2025personalized,wei2025personalized,yang2026longt2ibench,yang2025language}.

\begin{figure}[t]
	\centering
	\includegraphics[width=0.9\columnwidth]{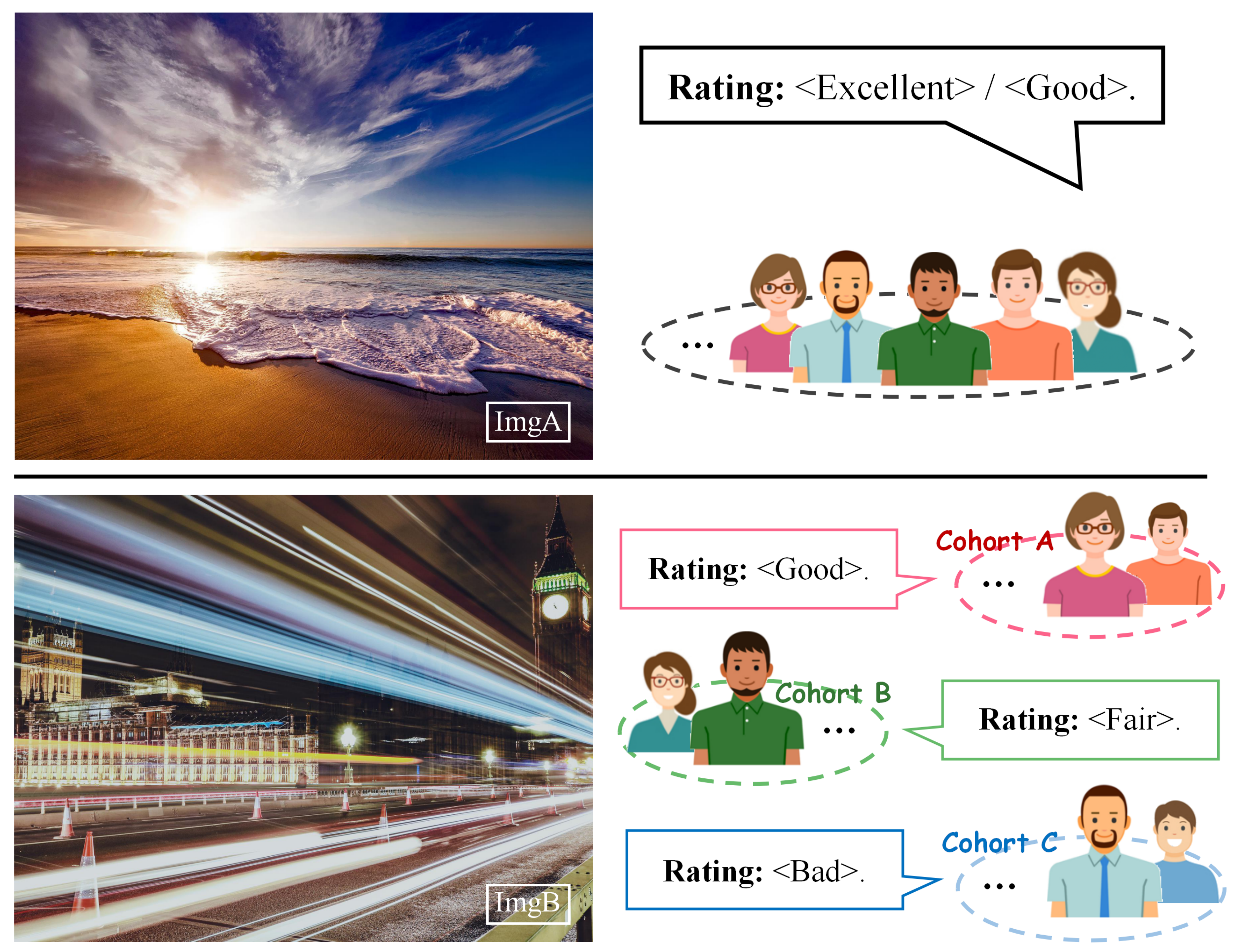} 
	\caption{Motivation of this study. Individual aesthetic differences vary in magnitude across distinct visual stimuli and exhibit cohort-specific distribution patterns. Two example images from the PARA database \protect\cite{yang2022personalized} are shown, along with their corresponding personalized aesthetic ratings.}
	\label{figb1}
\end{figure}

In the past few years, a proliferation of novel PIAA models have been reported in the literature. People's aesthetic tastes are typically determined by complex factors including both image attributes and user characteristics. This requires that models possess not only visual understanding capabilities but also knowledge of various demographic and personality dimensions, such as age, cultural background, and psychological traits \cite{kim2018objectivity}. To address these challenges, previous vision-based PIAA methods have largely resorted to auxiliary tasks related to preference representation, such as personalized feedback or personality trait prediction \cite{lv2021user,8970458}. Another challenge in PIAA is the few-shot learning (FSL) nature, as it is difficult to collect a large number of annotated images for a specific user. To address the dilemma, previous approaches are typically designed by fine-tuning generic IAA models \cite{ren2017personalized,wang2018collaborative}, or resort to meta-learning for extracting common aesthetic features across users that can efficiently transfer to target individuals \cite{zhu2020personalized,li2022transductive}. 

Despite the encouraging progress achieved, two important aspects remain underexplored in current PIAA methods. First, psychological research has demonstrated that some stimuli naturally elicit more active preference responses than others \cite{risko2012curious}, indicating that \textbf{not all images are equally informative for modeling aesthetic taste}. Second, researches in cognitive aesthetics \cite{celikors2025beauty} revealed that aesthetic preferences often cluster among individuals with shared sensibilities\textemdash`\textit{beauty is in the eye of your cohort}'\textemdash suggesting that \textbf{identifying aesthetically-resonant cohort offers valuable insights for target user modeling}. Inspired by the above psychological studies, we argue that 1) quantifying the richness of aesthetic preferences across images and 2) identifying aesthetically-resonant user cohorts, provide viable pathways for modeling personalized aesthetic preferences, as illustrated in Figure \ref{figb1}. In this context, identified informative samples and auxiliary cohorts can navigate the few-shot learning constraints in PIAA, maximizing the utility of the limited aesthetic annotations available for target users. With comprehensive visual understanding and extensive open-world knowledge of demographic and personality factors \cite{NEURIPS2023_21f7b745,li2023emotionprompt}, Multimodal Large Language Models (MLLMs) present natural advantages for sample measurement and cohort identification.

Motivated by the above facts, this paper presents \textbf{PRAC}, a novel MLLM-based PIAA model that captures individual aesthetics by preference-rich sample mining and aesthetically-resonant cohort merging. PRAC is designed with two key components: PreferSelect and PreferMerge. Specifically, the MLLM is first instructed to predict aesthetic divergence across the public through aesthetic distribution training on rich generic aesthetic data. Building upon the learned consensus aesthetic understanding and MLLMs' open-world knowledge, PreferSelect identifies preference-rich samples by analyzing both Collective Controversy and Personalized Deviation. \textit{Collective Controversy} measures the intrinsic aesthetic controversy of an image among the general public, while \textit{Personalized Deviation} captures the deviation of a specific user's aesthetic taste from the public consensus. Subsequently, targeted fine-tuning is performed on the identified preference-rich samples to encode user-specific aesthetic patterns. PreferMerge searches for aesthetically-resonant user cohorts by comparing MLLM-derived preference embeddings. Cohort-based model merging is then performed to aggregate preference patterns from aesthetically-resonant users with similar tastes to further enhance personalization for the target individual. The contributions of this work are three-fold:
\begin{itemize}
	\item A novel MLLM-based framework for personalized image aesthetic assessment (PRAC), which identifies informative samples and auxiliary cohorts to customize MLLMs for individual aesthetic tastes with limited annotations.
	\item We design two key components for preference modeling: PreferSelect mines preference-rich samples through dual-metric analysis of Collective Controversy and Personalized Deviation, while PreferMerge integrates preference patterns from aesthetically-resonant users via cohort-based merging.
	\item Extensive experiments conducted on four PIAA benchmarks demonstrate that PRAC outperforms state-of-the-arts on personalized rating prediction. Moreover, PRAC can offer rationales for varying aesthetic judgments, enhancing predictive interpretability.
\end{itemize}

\begin{figure*}
    \centering
    \includegraphics[width=0.92\textwidth]{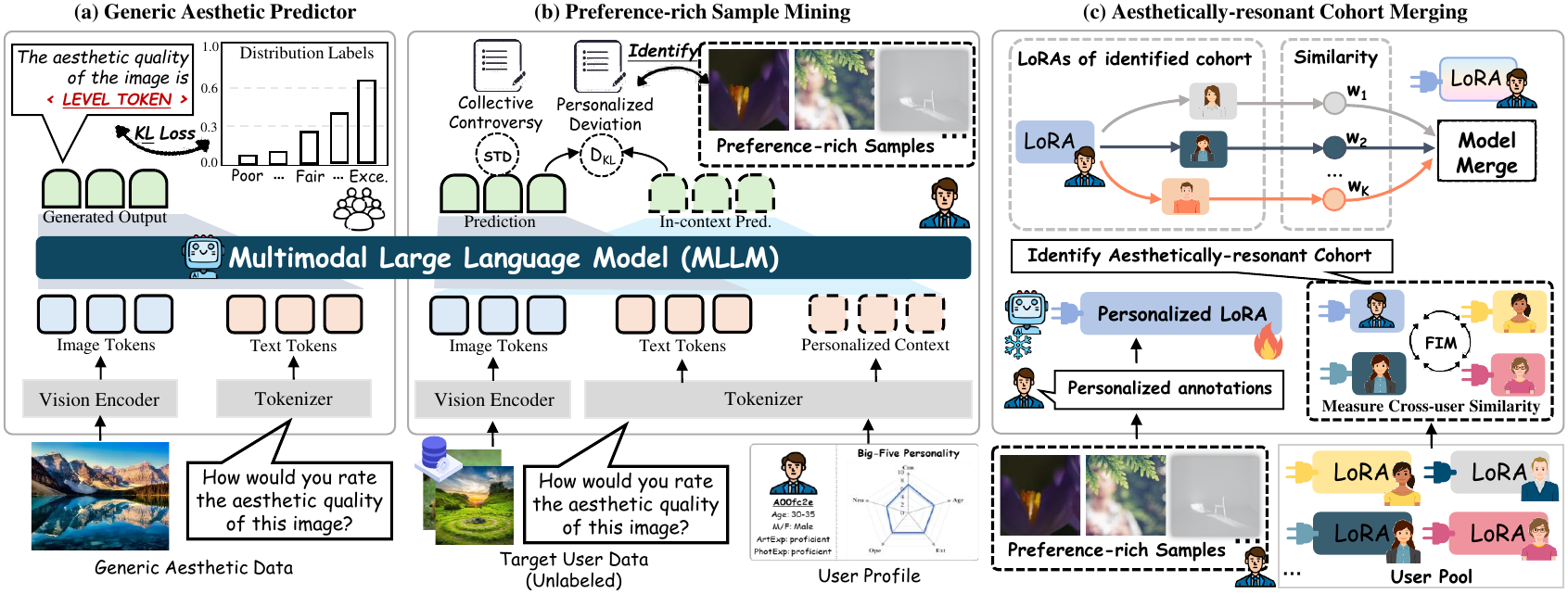}
    \caption{Overview of the proposed PRAC. The framework consists of three stages: (a) Generic Aesthetic Predictor: the MLLM is trained with aesthetic distribution supervision on generic aesthetic data to establish foundational aesthetic understanding; (b) PreferSelect: preference-rich samples are identified through collective controversy and personalized deviation metrics; (c) PreferMerge: the target user's personalized model is constructed by merging models from aesthetically-resonant cohort.}
    \label{fig2}
\end{figure*}

\section{Related Works}
\textbf{Personalized Image Aesthetic Assessment.} PIAA aims to model individual aesthetic tastes based on few user-provided aesthetic annotations. To capture aesthetic preferences, existing vision-based PIAA methods leveraged aesthetics-related auxiliary tasks, including users' photo favoriting behavior \cite{kim2018objectivity}, image re-ranking \cite{hamzah2012objectifying}, and personalized retouching \cite{lv2021user}, etc. Li \textit{et al.} found that individual aesthetic preference strongly correlates with their personality traits and proposed a multi-task learning framework to jointly estimate both personality traits and aesthetic preferences \cite{8970458}. Zhu \textit{et al.} further introduced a personalized aesthetic assessment approach that models aesthetic attributes of images as subjective factors and users' personality traits as objective factors \cite{zhu2021learning}. To address the few-shot learning challenge, meta-learning was adopted to identify common aesthetic features across users that can be adapted to target individuals \cite{li2022transductive,zhu2020personalized}. Recently, Yang \textit{et al.} applied contrastive learning to explore users' preference differences across various aesthetic levels \cite{10168279}. Yun \textit{et al.} innovatively employed Task Arithmetic to model target users' preferences by leveraging existing generic data \cite{yun2024scaling}. Different from these approaches, we model personalized aesthetics by quantifying preference richness across images and measuring preference similarity among users.

\textbf{LLMs/MLLMs-based Personalization.} The emergence of LLMs and MLLMs has opened new possibilities for personalization tasks due to their excellent capabilities in understanding complex patterns, reasoning about user behaviors, and generating contextually appropriate responses \cite{zhang2024personalization}. Recent efforts employ techniques like personalized prompting, adaptation, and alignment to tailor these models to individual users \cite{liu2025survey,huang2024aesexpert}. Tan \textit{et al.} employed personalized parameter-efficient fine-tuning to store user-specific behavior patterns and preferences, thereby democratizing LLMs \cite{tan2024democratizing}. Zhang \textit{et al.} proposed a parameterized memory-injected approach with a bayesian optimization search strategy to achieve LLM personalization \cite{zhang2024personalized}. In this study, we present the first exploration of personalizing MLLMs for modeling individual aesthetic preferences.

\textbf{Model Merging.} Model merging efficiently combines parameters from multiple specialized models to enhance overall capabilities without additional training. Representative approaches include linear interpolation, task arithmetic, and selective parameter merging\cite{li2023deep}. For LLMs and MLLMs, this paradigm enables capability integration without the prohibitive computational costs and data requirements of full-scale retraining \cite{yang2024model,lu2024twin}. Wu \textit{et al.} proposed aggregating parameters from lightweight task-specific experts learned from similar tasks to benefit target downstream tasks \cite{wu2023pi}. Tang \textit{et al.} designed a partial linearization method for parameter-efficient fine-tuned models, improving multi-task fusion capabilities with low computational overhead \cite{tang2023parameter}. In this paper, we introduce cohort-based model merging for PIAA task.

\section{Method}
In this section, we detail the proposed \textbf{PRAC}, which models aesthetic preferences by mining preference-rich samples and merging aesthetically-resonant cohorts. The overall framework is shown in Figure \ref{fig2}, which consists of three stages. First, we establish a Generic Aesthetic Predictor by training an MLLM on rich generic aesthetic data, enabling consensus aesthetic understanding. Building on this foundation and leveraging its powerful open-world knowledge, we design PreferSelect to mine preference-rich samples. Finally, we propose PreferMerge, a cohort-based model merging strategy that identifies aesthetically-resonant users and integrates them to further boost personalization for the target individual.

\subsection{Generic Aesthetic Predictor}

We first train a generic aesthetic predictor to model the distribution of public aesthetic judgments. Its ability to capture preference divergence serves as a crucial prerequisite for subsequent personalization stages. Given an image $I$, we obtain its aesthetic distribution from the statistics of individual annotations. The distribution label is represented as probabilities over five discrete aesthetic levels [Excellent, Good, Fair, Bad, Poor], denoted as $\mathbf{q} = [q_1, q_2, \dots, q_5]$, where $q_i$ denotes the probability of the $i$-th aesthetic level. Utilizing these generic aesthetic data, we integrate a parameter-efficient LoRA module \cite{hu2022lora} into MLLMs to conduct aesthetic distribution training. Specifically, we design specialized prompt templates to guide MLLMs in generating aesthetic distributions and employ the Kullback-Leibler (KL) Divergence loss to optimize parameters.

\textbf{Guiding MLLMs for Distribution Prediction.} Inspired by the training of previous MLLMs-based scoring methods \cite{wu2024q,you2025teaching}, we prompt the MLLM with an image along with a fixed question: `\textit{How would you rate the aesthetic quality of this image?}'. Simultaneously, we specify the output template as `\textit{The aesthetic quality of the image is $<$Level Token$>$}', where \textit{$<$Level Token$>$} represents the predicted aesthetic level. Given that the token constitutes probabilities over all possible vocabulary in the language model, we perform a close-set softmax over the five textual aesthetic levels to obtain predicted aesthetic distribution of MLLMs, denoted as $\mathbf{p} = [p_1, p_2, \dots, p_5]$.

\textbf{KL Loss for Parameter Update.} For the $i$-th image $I^i$, we compute the KL divergence between the predicted distribution $\mathbf{p}^i $ from the MLLM and the ground-truth distribution $\mathbf{q}^i$ as the training loss. By aggregating the total loss $\mathcal{L}_{KL}$ across the entire training set, we optimize the parameters of the integrated LoRA module. This is formulated as:
\begin{equation}
	\begin{split}
		\mathcal{L}_{KL} &= \frac{1}{N_g} \sum_{i=1}^{N_g} D_{KL}(\mathbf{q}^i \parallel \mathbf{p}^i) \\
		&= \frac{1}{N_g} \sum_{i=1}^{N_g} \sum_{j=1}^{5} q_j^i \log\left(\frac{q_j^i}{p_j^i}\right),
	\end{split}
\end{equation}
where $D_{KL}$ denotes the KL divergence operation, $ N_g $ is the number of training images and $j$ indexes the five aesthetic levels of [Excellent, Good, Fair, Bad, Poor].

\begin{figure}[t]
	\centering
	\includegraphics[width=0.99\columnwidth]{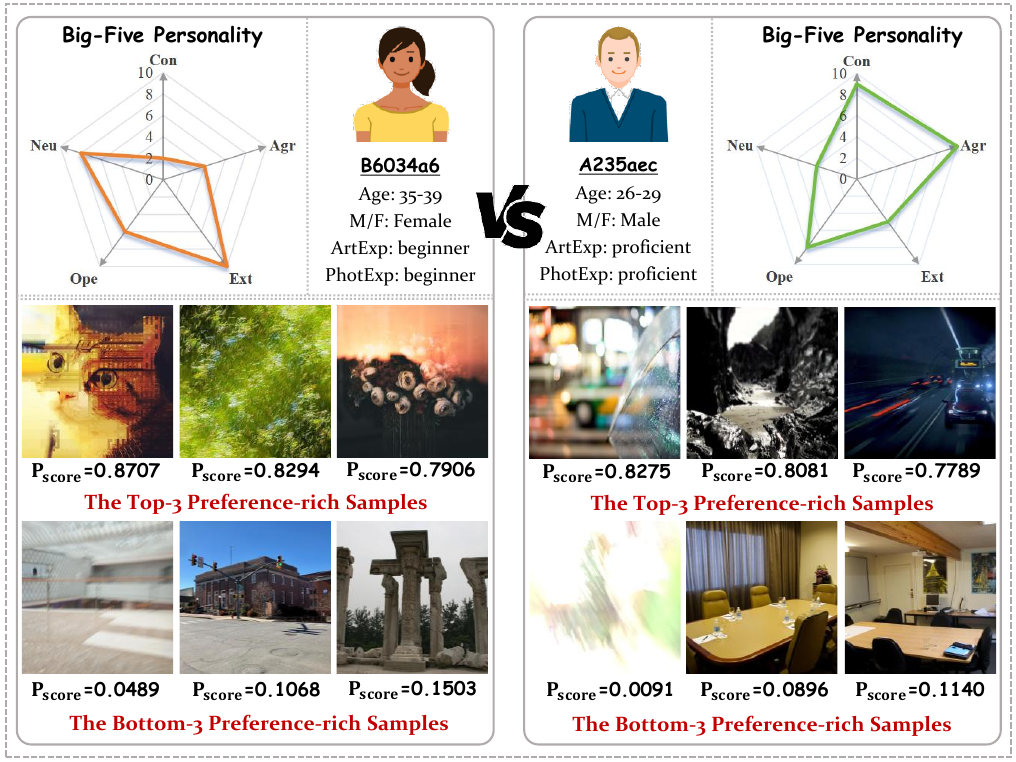} 
	\caption{Visualization of PreferSelect results. Two users from the PARA database \protect\cite{yang2022personalized} with distinct profiles (demographics and Big-Five personality traits) and their corresponding top-3/bottom-3 preference-rich samples based on $\text{P}_{\text{score}}$ rankings.}
	\label{fig3}
\end{figure}

\subsection{Preference-rich Sample Mining}

The varying richness of aesthetic preference across images lies in the fact that certain samples are more informative for personalization. PreferSelect aims to utilize the acquired aesthetics principles of the Generic Aesthetic Predictor along with MLLM's extensive open-world knowledge to identify preference-rich samples. Inspired by research in cognitive aesthetics showing that aesthetic experiences arise from the interplay of demographics and personality traits \cite{washizu2025bodily}, we design a dual-metric selection mechanism to measure the preference richness of samples. This mechanism synergistically combines two perspectives: 1) the intrinsic aesthetic controversy of an image among the general public (Collective Controversy), and 2) the predicted deviation of a specific user's taste from the public (Personalized Deviation).

\textbf{Collective Controversy Metric ($\text{CCM}$).} Images eliciting significant controversy in aesthetic judgments inherently encode richer personalized preference information. Based on this observation, we design a simple yet effective metric to quantify such controversy. Specifically, for a given image $I^i$, we utilize the Generic Aesthetics Predictor to predict its aesthetic distribution $\mathbf{p}^i $ and calculate the standard deviation (STD) of this distribution as the measurement:
\begin{equation}
	\text{CCM}(I^i)= \sigma (\mathbf{p}^i) = \sqrt{\sum_{j=1}^{5} (j - \mu^i   )^2 \cdot {p_j^i}},
\end{equation}
where $\sigma$ denotes the standard deviation, and $\mu^i=\sum_{j=1}^{5}j\cdot p_{j}^{i}$ is the mean of the distribution for image $I^i$.

\textbf{Personalized Deviation Metric ($\text{PDM}$).} People's aesthetic preferences are significantly influenced by their intrinsic attributes, particularly personality traits and photographic expertise \cite{yang2022personalized}. MLLMs with extensive open-world knowledge of demographic and personality traits \cite{NEURIPS2023_21f7b745}, provide a foundation for measuring the impact of these factors on aesthetic judgments. Inspired by in-context learning \cite{dong2024survey}, we convert discrete user profiles (age, traits, expertise) into natural language descriptions and integrate them into the prompt: `\textit{You are a $<$User Profile$>$, how would you rate the aesthetic quality of this image?}'. This prompt is then input to the Generic Aesthetic Predictor to generate an in-context aesthetic distribution. The PDM is defined as the KL Divergence between the generic distribution $\mathbf{p}^i$ and the in-context distribution $\mathbf{p}_u^i$:
\begin{equation}
	\text{PDM}(I^i) = D_{KL}(\mathbf{p}_u^i \parallel \mathbf{p}^i). 
\end{equation}

To comprehensively integrate both Collective Controversy and Personalized Deviation, we propose a unified preference richness score $\text{P}_{\text{score}}$. For image $I^i$, it is formulated as a weighted combination of the two fundamental metrics:
\begin{equation}
	\text{P}_{\text{score}}(I^i) = (1-\alpha) \cdot \text{CCM}(I^i) + \alpha \cdot \text{PDM}(I^i), 
\end{equation}
where $\alpha$ is a weighting hyperparameter that balances the contributions of collective controversy and personalized deviation. For a specific user $U_i$ with his/her personal images, we compute $\text{P}_{\text{score}}$ for each image and rank all candidates accordingly. Images with higher scores are selected to form the personalized query set $\mathcal D_{u}$ for the user. This curated set facilitates efficient encoding of user-specific aesthetic patterns in the PreferMerge stage. Figure \ref{fig3} illustrates the PreferSelect results: two users with distinct demographic and personality traits demonstrate markedly different preference-rich samples. Furthermore, within each user, top-3 samples exhibit significantly higher $\text{P}_{\text{score}}$ than bottom-3 samples, validating the quantification of preference-richness.

\subsection{Aesthetically-resonant Cohort Merging}

Based on the identified preference-rich images, we propose PreferMerge, a cohort-based model merging strategy that discovers and integrates aesthetically-resonant user cohorts to enhance target user personalization. Under limited annotation scenarios (e.g., 10-shot or 100-shot settings, commonly adopted in the PIAA task) where the target user provides aesthetic ratings on preference-rich samples, PreferMerge first measures aesthetic preference similarity across users, then identifies and merges aesthetically-resonant users to strengthen personalization for the target individual.

\textbf{Measuring Cross-user Preference Similarity.} Individual annotations on preference-rich samples directly reflect their aesthetic sensibilities, providing a foundation for measuring cross-user preference similarity. We fine-tune the Generic Aesthetic Predictor using these personalized annotations and extract gradient directions as preference embeddings. The correlation between preference embeddings serves as a robust metric for quantifying preference similarity among users.

Given a query set $\mathcal D_u={(I^k, y^k) _{k=1}^{N_u}}$ of preference-rich samples for user $U_i$, where $ y^k $ denotes the user's personalized annotation to image $ I^k $ as a discrete aesthetic level label, we perform targeted fine-tuning. A user-specific LoRA module is injected into the Generic Aesthetic Predictor and fine-tuned exclusively on its parameters $\theta_u$. The cross-entropy loss between predicted and ground-truth aesthetic levels guides the parameter updates. We compute the gradient fluctuations of parameters $\theta_u$ on the query set $\mathcal D_u$ as the preference embedding for user $U_i$. Specifically, we employ the Fisher Information Matrix (FIM) as an indicator \cite{amari1998natural}, which quantifies the model's sensitivity to parameter perturbations under limited data:
\begin{equation}
	\mathbf{F}_{u}=\mathop{\mathbb E}\limits_{(I, y) \sim \mathcal{D}_{u}}\left[\nabla_{\theta_{u}} \log p\left(y \mid I\right) \nabla_{\theta_{u}} \log p\left(y \mid I\right)^{T}\right],
\end{equation}
where $\mathbf{F}_{u}$ represents the preference embedding. For two users $U_i$ and $U_j$, with respective embeddings $\mathbf{F}_{u}^{i}$ and $\mathbf{F}_{u}^{j}$, we compute the cosine distance between them to measure preference similarity $	\operatorname{Sim}\left(U_{i}, U_{j}\right)$:
\begin{equation}
	\operatorname{Sim}\left(U_{i}, U_{j}\right)=\frac{\mathbf{F}_{u}^{i}  \cdot \mathbf{F}_{u}^{j}}{\left\|\mathbf{F}_{u}^{i}\right\| \cdot \left\|\mathbf{F}_{u}^{j}\right\|}.
\end{equation}

\textbf{Cohort Selection and Merging.} With cross-user preference similarities quantified, we identify an optimal cohort of aesthetically-resonant users and merge them to enhance target user personalization. The selection is performed on a pre-constructed User Pool, which contains personalized LoRAs from multiple training users obtained through targeted fine-tuning on their respective query sets, serving as a rich library of diverse aesthetic preferences. The selection strategy balances two key objectives: Target Relevance and Cohort Diversity. Target Relevance ensures that selected users are maximally relevant to the target individual, while Cohort Diversity prevents redundancy among the selected users. To balance these two aspects, we introduce a trade-off parameter $\beta $ and formulate the optimal selection problem as:
\begin{equation}
	\begin{aligned}
		\mathcal S^{*} = \mathop{\arg\max}\limits_{\mathcal S\subset \mathcal P,\left |\mathcal S \right |=K} \Big[ & \beta \cdot \sum_{U_{i}\in \mathcal S} \operatorname{Sim}\left(U_{target}, U_{i}\right) - \\
		& (1-\beta)  \cdot \sum_{U_{i},U_{j}\in \mathcal S, i \neq j} \operatorname{Sim}\left(U_{i}, U_{j}\right) \Big],
	\end{aligned}
\end{equation}
where $\mathcal P$ denotes the User Pool, and $\mathcal S^{*} $ is the optimal subset of $K$ aesthetically-resonant users that maximizes relevance to the target user while minimizing intra-cohort redundancy. 

Once the optimal set $\mathcal S^{*}$ is identified, we merge the personalized LoRA models from these selected users through a weighted fusion operation. Each model contributes proportionally to its preference similarity with the target user:
\begin{equation}
	\theta_{merged} = \sum_{i=1}^{K} w_i \cdot \theta _{u}^{i},
\end{equation}
where $ w_i $ represents the normalized merging weight for model $\theta _{u}^{i} $. This targeted merging efficiently leverages preference patterns from users with similar aesthetic tastes to enhance the personalization for the target individual.

\section{Experiments}
\subsection{Experimental Settings}

\textbf{Databases.} To evaluate the proposed PRAC, we conduct extensive experiments on four PIAA benchmarks: PARA \cite{yang2022personalized}, FLICKR-AES \cite{ren2017personalized}, REAL-CUR \cite{ren2017personalized}, and AADB \cite{kong2016photo}. In these datasets, each image is annotated by multiple users for preference modeling, with key statistics summarized in Table \ref{tab1}. Specifically, \textit{Ann. Users per Img} and \textit{Ann. Imgs per User} denote the average number of users annotating each image and the average number of images annotated per user, respectively. Notably, PARA \cite{yang2022personalized} provides rich user profiles characterized by six attributes: age, gender, education, artistic experience, photographic experience, and personality traits. Moreover, REAL-CUR \cite{ren2017personalized} consists of photographs from real personal albums, where each image is rated exclusively by its owner.

\begin{table}[h]
	\centering
	\caption{Statistics of PIAA datasets. `/' denotes train/test split.}
	\label{tab1}
	
	\resizebox{\linewidth}{!}
	{
		\begin{tabular}{l|c|c|c|c}
			\toprule
			\textbf{Database} & \textbf{PARA} & \textbf{FLICKR-AES} & \textbf{REAL-CUR} & \textbf{AADB} \\
			\midrule
			\midrule
			Num. of Imgs & 28,220 / 3,000 & 35,263 / 4,737 & 0 / 2,871 & 8,458 / 1,542 \\
			Num. of Users  & 398 / 40 & 173 / 37 & 0 / 14 & 168 / 22 \\
			Ann. Users per Img & 25 & 5 & 1 & 5 \\
			Ann. Imgs per User & 140--3,500 & 105--171 & 197--222 & 110--190 \\
			\bottomrule
		\end{tabular}
	}
\end{table}

\textbf{Implementation Details.} We adopt the open-source MLLM mPLUG-Owl3 \cite{ye2024mplug} as our base model.  During the Generic Aesthetic Predictor phase, images labeled with distribution statistics derived from multiple users' annotations within the target PIAA databases are employed for training. We construct the User Pool using training user data from each dataset, while the testing users serve as targets. The hyperparameters are set to $\alpha$=0.3, $K$=6, and $\beta$=0.5, with detailed ablation studies provided in Section \ref{sec:ablation}. Notably, both the training and inference of PRAC for a target user can be executed on a single NVIDIA RTX 3090 GPU, with memory consumption and inference speed detailed in Section \ref{sec:cost}.

\textbf{Evaluation Criterion.} Similar to the previous literature, we employ the Spearman Rank-order Correlation Coefficient (SRCC) as a criterion to validate the performance of PIAA models. Furthermore, there are two commonly used testing patterns based on the number of images in the query set $\mathcal D_u$, namely 10-shot and 100-shot \cite{ren2017personalized}.

\begin{table}[t]
	\centering
    \caption{Comparison between PRAC and state-of-the-art PIAA methods, with the average SRCC of 40 testing users from PARA \protect\cite{yang2022personalized} in 10-shot and 100-shot testing modes.}
	\label{tab2}
	\resizebox{0.8\linewidth}{!}
	{
		\begin{tabular}{c cc}
			\toprule
			\textbf{Methods} & \textbf{10-shot} & \textbf{100-shot} \\
			\midrule
			\midrule
			PARA (unconditional)   & 0.681 & 0.695 \\
			PARA (artistic)     & 0.686 & 0.698 \\
			PARA (photographic) & 0.683 & 0.698 \\
			PARA (personality)  & 0.691 & 0.705 \\
			\midrule
			BLG-PIAA \cite{zhu2020personalized}            & 0.688 & 0.698 \\
			PA-IAA \cite{8970458}              & 0.683 & 0.696 \\
			PIAA-SOA \cite{zhu2021learning}            & 0.690 & 0.703 \\
			PIAA-MIR \cite{zhu2022personalized}            & 0.702 & 0.716 \\
			MTCL \cite{10168279}                & 0.695 & 0.713 \\
			\midrule
			\textbf{PRAC (Ours)} & \textbf{0.707} & \textbf{0.733} \\
			\bottomrule
		\end{tabular}
	}
\end{table}

\begin{table}[t]
	\centering
    \caption{Comparison between PRAC and 11 representative PIAA models across the 37 testing users of the FLICKR-AES database \protect\cite{ren2017personalized}. The best results are shown in bold.}
	\label{tab3}
	\resizebox{0.8\linewidth}{!}
	{
		\begin{tabular}{c cc}
			\toprule
			\textbf{Methods} & \textbf{10-shot} & \textbf{100-shot} \\
			\midrule
			\midrule
			PAM (attribute only) & 0.511 & 0.516 \\
			PAM (content only) & 0.512 & 0.516 \\
			PAM \cite{ren2017personalized} & 0.513 & 0.524 \\
			\midrule
			USAR-PPR & 0.521 & 0.544 \\
			USAR-PAD & 0.520 & 0.537 \\
			USAR \cite{lv2018usar} & 0.525 & 0.552 \\
			\midrule
			PASS \cite{wang2018collaborative} & 0.516 & 0.521 \\
			PA-IAA \cite{8970458} & 0.543 & 0.639 \\
			BLG-PIAA \cite{zhu2020personalized} & 0.561 & 0.669 \\
			UG-PIAA \cite{lv2021user} & 0.559 & 0.660 \\
			PIAA-SOA \cite{zhu2021learning}  & 0.618 & 0.691 \\
			TAPP-PIAA \cite{li2022transductive} & 0.591 & 0.685 \\
			IM-PIAA \cite{hou2022interaction} & 0.620 & 0.708 \\
			MTCL \cite{10168279} & 0.667 & 0.737 \\
			Yun \textit{et al.} \cite{yun2024scaling} & 0.668 & 0.748 \\
			\midrule
			\textbf{PRAC (Ours)} & \textbf{0.692} & \textbf{0.778} \\
			\bottomrule
		\end{tabular}
	}
\end{table}

\begin{table}[t]
	\centering
    \caption{Performance comparison conducted on 22 testing users from the AADB \protect\cite{kong2016photo} database.}
	\label{tab4}
	\resizebox{0.8\linewidth}{!}
	{
	\begin{tabular}{c cc}
		\toprule
		\textbf{Methods} & \textbf{10-shot} & \textbf{100-shot} \\
		\midrule
		\midrule
		BA-PIAA \cite{zhu2020personalized} & 0.450 & 0.513 \\
		BLG-PIAA \cite{zhu2020personalized} & 0.497 & 0.545 \\
		Inductive-PIAA \cite{li2022transductive} & 0.524 & 0.565 \\
		TAPP-PIAA \cite{li2022transductive} & 0.534 & 0.612 \\
		MTCL \cite{10168279} & 0.540 & 0.622 \\
		Yun \textit{et al}. \cite{yun2024scaling} & 0.556 & 0.654 \\
		\midrule
		\textbf{PRAC (Ours)} & \textbf{0.597} & \textbf{0.671} \\
		\bottomrule
	\end{tabular}
	}
\end{table}

\begin{table}[t]
	\centering
    \caption{Cross-database evaluation. Models trained on FLICKR-AES \protect\cite{ren2017personalized} and evaluated on testing users from REAL-CUR \protect\cite{ren2017personalized} and AADB \protect\cite{kong2016photo} databases.}
	\label{tab5}
	\resizebox{0.9\linewidth}{!}
	{
	\begin{tabular}{l|l|c|c}
		\toprule
		\textbf{Databases} & \textbf{Methods} & \textbf{10-shot} & \textbf{100-shot} \\
		\midrule
		\midrule
		\multirow{7}{*}{REAL-CUR} 
		& PA-PIAA \cite{8970458} & 0.443 & 0.562 \\
		& BLG-PIAA \cite{zhu2020personalized} & 0.448 & 0.578 \\
		& PIAA-SOA \cite{zhu2021learning} & 0.487 & 0.589 \\
		& TAPP-PIAA \cite{li2022transductive} & - & 0.580 \\
		& MTCL \cite{10168279} & 0.495 & 0.599 \\
		& Yun \textit{et al}. \cite{yun2024scaling} & 0.577 & 0.621 \\
		& \textbf{PRAC (Ours)} & \textbf{0.585} & \textbf{0.631} \\
		
		\midrule
		
		\multirow{6}{*}{AADB} 
		& PA-PIAA \cite{8970458} & 0.469 & 0.524 \\
		& BLG-PIAA \cite{zhu2020personalized} & 0.486 & 0.536 \\
		& PIAA-SOA \cite{zhu2021learning} & 0.509 & 0.557 \\
		& TAPP-PIAA \cite{li2022transductive} & - & 0.540 \\
		& MTCL \cite{10168279} & 0.533 & 0.572 \\
		& \textbf{PRAC (Ours)} & \textbf{0.547} & \textbf{0.589} \\
		\bottomrule
	\end{tabular}
	}
\end{table}

\subsection{Performance Evaluation}

\textbf{Intra-database Evaluation.} We first evaluate the performance of the proposed PRAC in modeling personalized aesthetic preferences on the PARA \cite{yang2022personalized}, FLICKR-AES \cite{ren2017personalized}, and AADB \cite{kong2016photo} databases. The experimental results are presented in Table \ref{tab2}, Table \ref{tab3}, and Table \ref{tab4}, respectively. PRAC consistently achieves the best performance across all datasets, demonstrating that PreferSelect and PreferMerge accurately capture individual aesthetic preferences. Notably, compared to auxiliary task-based methods such as USAR (user-specific re-ranking) \cite{lv2018usar}, UG-PIAA (user personalized retouching) \cite{lv2021user}, and IM-PIAA (user content preference) \cite{hou2022interaction}, PRAC achieves superior results using only aesthetic annotations, highlighting the advantages of leveraging MLLMs for preference modeling. Furthermore, PRAC outperforms meta-learning based methods such as BLG-PIAA \cite{zhu2020personalized} and TAPP-PIAA \cite{li2022transductive}, validating the effectiveness of preference-rich sample mining and aesthetically-resonant cohort merging for few-shot preference modeling.

\textbf{Cross-database Evaluation.} To verify the generalization performance of the PRAC, we further conduct a cross-database testing experiment. Following the protocol established in previous work \cite{8970458,zhu2020personalized,10168279}, we utilized the model trained on the FLICKR-AES dataset \cite{ren2017personalized} to evaluate the personalization performance on 14 real users from REAL-CUR \cite{ren2017personalized} and 22 testing users from AADB \cite{kong2016photo}. Table \ref{tab5} presents the comparative results, where PRAC consistently achieves state-of-the-art performance. These results demonstrate that the proposed PRAC has a strong generalization ability, which is important for real world applications.

\begin{table}[t]
	\centering
    \caption{Ablation study of PRAC components across the four PIAA databases. The \checkmark indicates the component is enabled.}
	\label{tab6}
	\resizebox{0.99\linewidth}{!}
	{
	\begin{tabular}{c c c c c}
		\toprule
		\textbf{Databases} & \textbf{PreferSelect} & \textbf{PreferMerge} & \textbf{10-shot} & \textbf{100-shot} \\
		\midrule
		\midrule
		\multirow{3}{*}{PARA} 
		& -- & -- & 0.634 & 0.672 \\
		& \checkmark & -- & 0.640 & 0.701 \\
		& \checkmark & \checkmark & 0.707 & 0.733 \\
		\midrule
		\multirow{3}{*}{FLICKR-AES} 
		& -- & -- & 0.644 & 0.672 \\
		& \checkmark & -- & 0.661 & 0.717 \\
		& \checkmark & \checkmark & 0.692 & 0.778 \\
		\midrule
		\multirow{3}{*}{AADB} 
		& -- & -- & 0.541 & 0.602 \\
		& \checkmark & -- & 0.570 & 0.613 \\
		& \checkmark & \checkmark & 0.597 & 0.671 \\
		\midrule
		\multirow{3}{*}{REAL-CUR} 
		& -- & -- & 0.495 & 0.524 \\
		& \checkmark & -- & 0.504 & 0.560 \\
		& \checkmark & \checkmark & 0.585 & 0.631 \\
		\bottomrule
	\end{tabular}
	}
\end{table}

\begin{table}[t]
	\centering
    \caption{Personalization performance of different open-source MLLMs with and without the proposed PRAC on the PIAA task evaluated on PARA database.}
	\label{tab7}
	\resizebox{0.9\linewidth}{!}
	{
	\begin{tabular}{lcc}
		\toprule
		\textbf{Backbone} & \textbf{10-shot} & \textbf{100-shot} \\
		\midrule
		\midrule
		Qwen3-VL (w/o PRAC) & 0.540 & 0.574 \\
		Qwen3-VL (w PRAC) & 0.694 & 0.723 \\
		\midrule
		InternVL-3.5 (w/o PRAC) & 0.394 & 0.478 \\
		InternVL-3.5 (w PRAC) & 0.690 & 0.718 \\
		\midrule
		mPLUG-Owl3 (w/o PRAC) & 0.497 & 0.647 \\
		mPLUG-Owl3 (w PRAC) & 0.707 & 0.733 \\
		\bottomrule
	\end{tabular}
	}
\end{table}

\subsection{Ablation Study}
\label{sec:ablation}

\textbf{Contributions of Model Components.} To validate the effectiveness of each component in PRAC, we conduct comprehensive ablation experiments on the four benchmark PIAA datasets, as summarized in Table \ref{tab6}. First, we examine the personalization performance of the Generic Aesthetic Predictor (baseline) fine-tuned directly on randomly selected user query sets without PreferSelect and PreferMerge. To ensure unbiased evaluation, experiments for all testing users were conducted 10 times with random sampling of training images, and the average results were recorded. The results clearly demonstrate that PRAC outperforms the Generic Aesthetic Predictor by substantial margins, with performance improvements exceeding 5\% across all datasets in both 10-shot and 100-shot settings. Furthermore, we test the model using only PreferSelect, which also shows notable improvements over the baseline. These results provide evidence for the effectiveness of both PreferSelect and PreferMerge in personalized aesthetic modeling.

\textbf{Impact of MLLMs.} We compare the personalization performance of popular open-source MLLMs (Qwen3-VL \cite{bai2023qwen}, InternVL-3.5 \cite{wang2025internvl3}, and mPLUG-Owl3 \cite{ye2024mplug}) with and without PRAC on the PARA dataset, as illustrated in Table \ref{tab7}. The results demonstrate that PRAC consistently outperforms direct fine-tuning (no personalization) in both 10-shot and 100-shot settings across all MLLMs. Moreover, the stable performance gains validate that PRAC is model-agnostic and can be effectively integrated with various MLLM architectures to enhance personalized aesthetic modeling.

\begin{figure}[t]
	\centering
	\includegraphics[width=0.97\columnwidth]{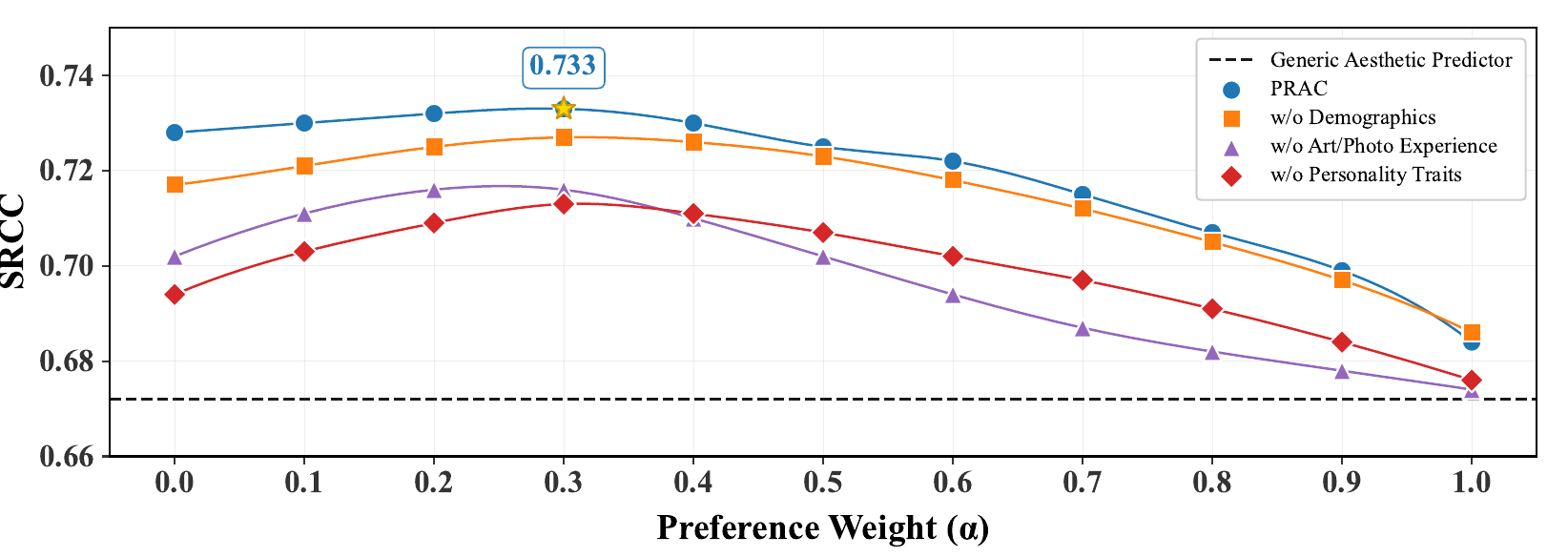} 
	\caption{Ablation study on weight $\alpha$ (balancing CCM and PDM) and user profiles (demographics, art/photography experience, and personality traits) in PreferSelect.}
	\label{fig4}
\end{figure}

\begin{figure}[t]
	\centering
	\includegraphics[width=0.97\columnwidth]{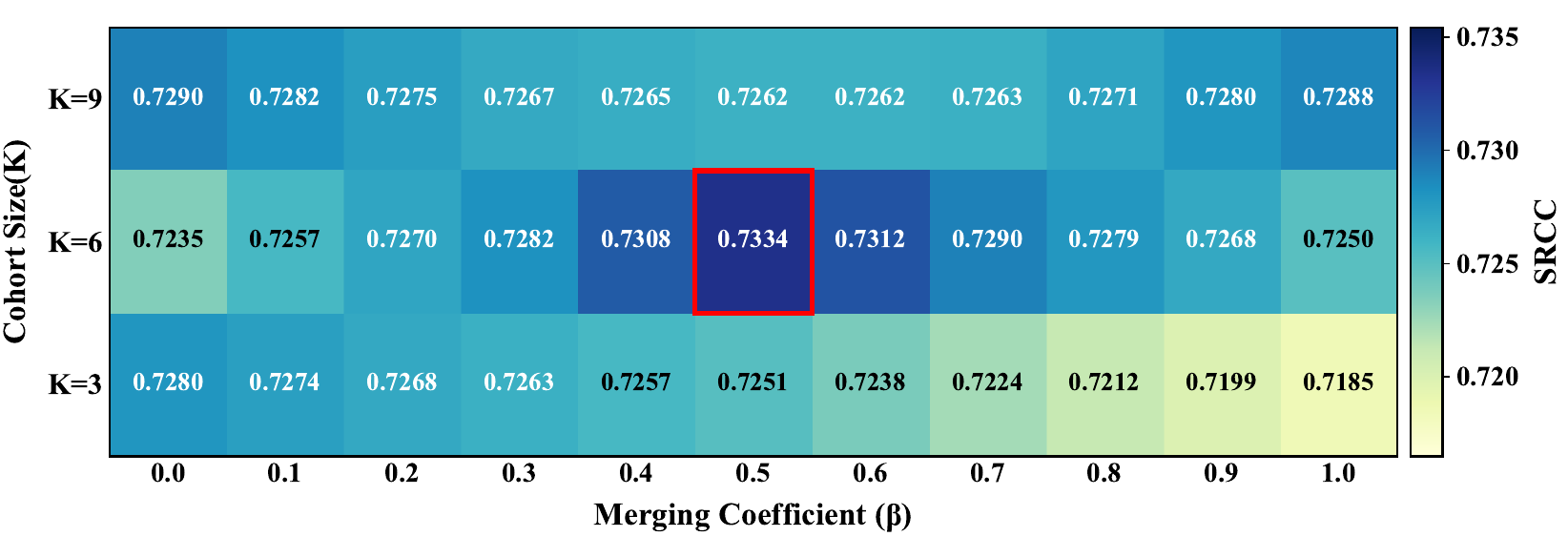} 
	\caption{Ablation study on aesthetically-resonant cohort size $K$ and  the hyperparameter $\beta$ (balancing Target Relevance and Cohort Diversity) in PreferMerge.}
	\label{fig5}
\end{figure}

\textbf{Ablation of PreferSelect.} The hyperparameter $\alpha $ (balancing CCM and PDM) and user profiles (demographics, art/photography experience, and personality traits), jointly influence preference-rich sample mining in PreferSelect. To scrutinize these effects, we conduct an ablation study on the PARA database \cite{yang2022personalized}, evaluating how PreferSelect influences the modeling of personalized aesthetic preferences under varying balancing weights and when different types of user information are excluded, as illustrated in Figure \ref{fig4}. Results show that $\alpha=0.3$ achieves optimal performance, and all ablation settings outperform the baseline (w/o PreferSelect), validating the importance of preference-rich sample mining. Moreover, user traits notably impact the measurement of preference richness, underscoring the critical role of personality traits in shaping aesthetic preferences, as similarly evidenced in PIAA studies \cite{8970458, zhu2021learning}.

\textbf{Ablation of PreferMerge.} The search for aesthetically-resonant users and the cohort size used for merging are crucial factors affecting the prediction of personalized aesthetic preferences for target individuals. Figure \ref{fig5} presents an ablation study on the hyperparameter $\beta$ (balancing Target Relevance and Cohort Diversity) and the cohort size $K$ in PreferMerge. Optimal performance is achieved at $K=6, \beta=0.5$. The results reveal that smaller cohorts ($K=3$) benefit from higher diversity weighting, while larger cohorts ($K=9$) suffer from user redundancy, leading to performance degradation. The setting of $\beta=0.5$ demonstrates that the optimal aesthetically-resonant cohort takes into account both relevance and diversity.

\subsection{Qualitative Analysis}
\begin{figure}[t]
	\centering
	\includegraphics[width=0.97\columnwidth]{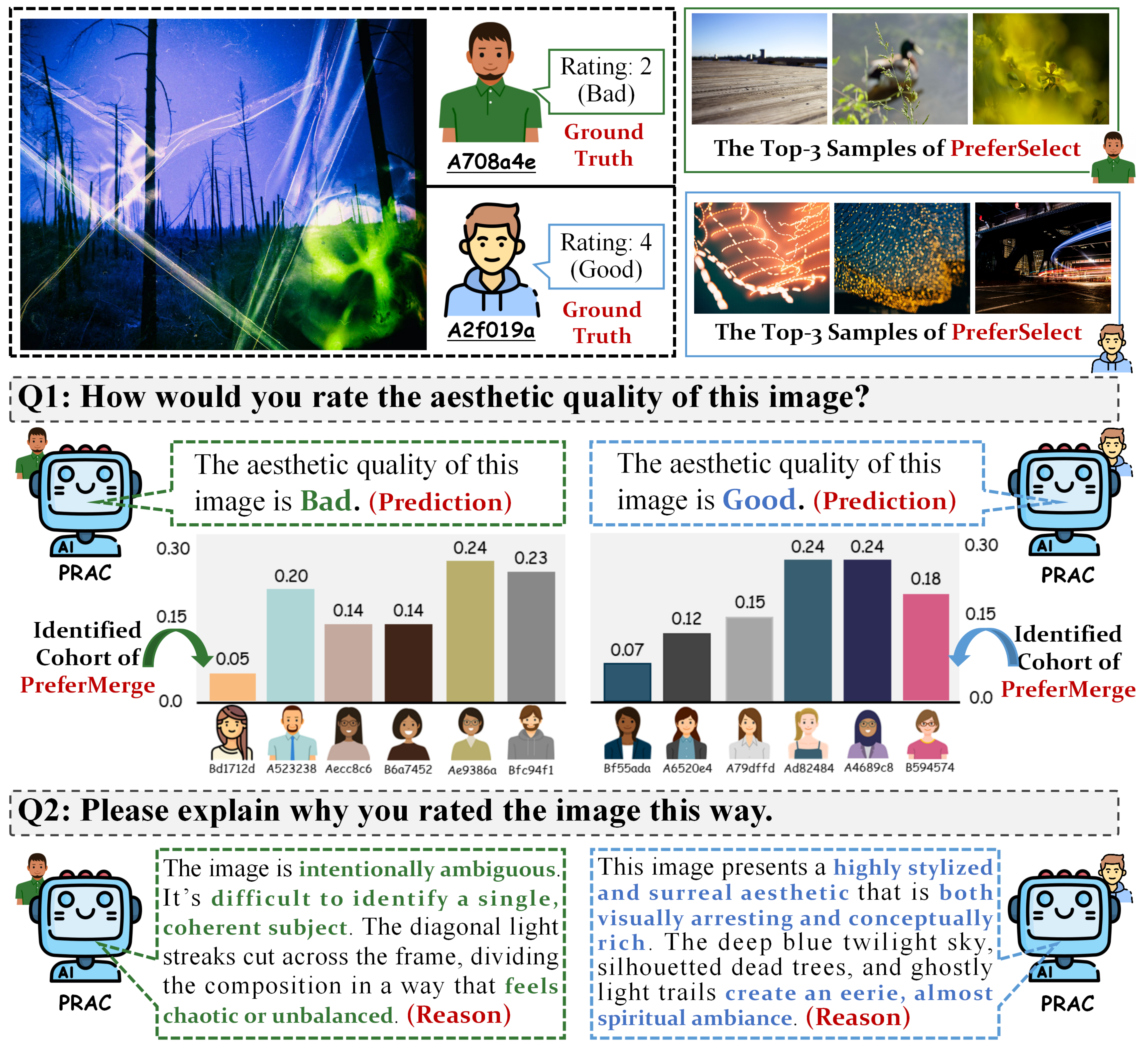} 
	\caption{Visualization of personalization modeling in PRAC with two testing users from the PARA database \cite{yang2022personalized}.}
	\label{fig6}
\end{figure}

\textbf{How PRAC Personalizes Using PreferSelect and PreferMerge?} Beyond quantitative evaluation on rating prediction, we conduct a qualitative study using two example users to visualize how the proposed PRAC model performs sample mining and cohort merging, ultimately facilitating personalized aesthetic expression. As illustrated in Figure \ref{fig6}, for each testing user, we visualize: (1) the top-3 preference-rich samples mined by PreferSelect; (2) the identified cohort with merging weights from PreferMerge; (3) predicted ratings and reasons generated using the prompt `Please explain why you rated the image this way'. The two users exhibit opposite evaluations (Bad vs Good) on the same image, supported by distinct explanations, different samples from PreferSelect, and diverse cohorts from PreferMerge. This demonstrates the effectiveness of PRAC in capturing personalized aesthetic preferences.

\begin{figure}[t]
	\centering
	\includegraphics[width=0.9\columnwidth]{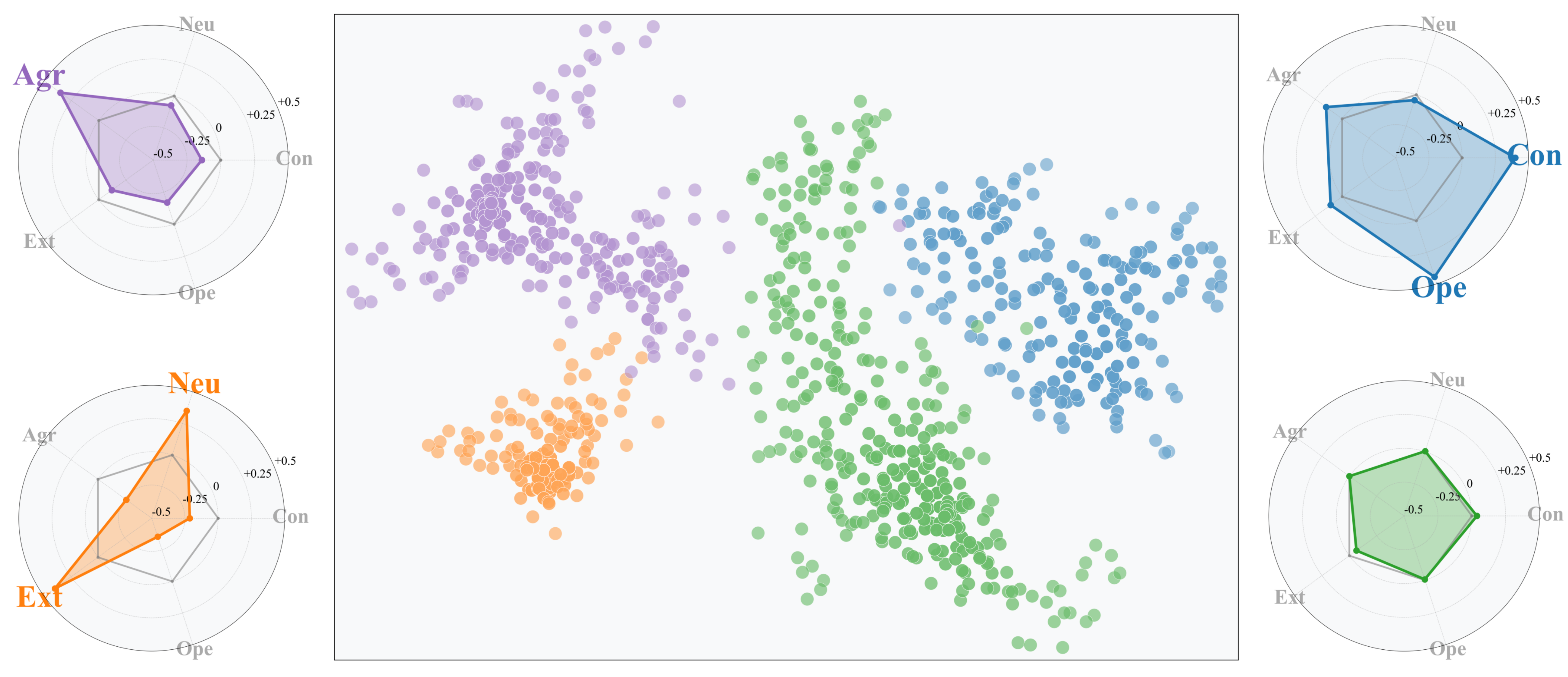} 
	\caption{Clustering distribution of 398 users based on their Big-Five personality traits and Preference Embeddings.}
	\label{figb2}
\end{figure}

\textbf{How Do Personality Traits relate to Preference Embeddings?} Personality traits play a crucial role in preference-rich sample mining, while preference embeddings facilitate aesthetically-resonant cohort merging. Preference embeddings are derived by calculating parameter perturbations during the fine-tuning of the MLLM on selected preference-rich samples. To investigate the relationship between the Big-Five Personality Traits and Preference Embeddings, we conduct a qualitative analysis involving 398 users in the PARA \cite{yang2022personalized}. Specifically, we recorded each user's preference embedding alongside their personality traits, which include five dimensions: Openness, Conscientiousness, Extraversion, Agreeableness, and Neuroticism. Subsequently, unsupervised clustering was applied across all users, followed by the calculation of average personality traits for each cluster, as illustrated in Figure \ref{figb2}. The results indicate that users exhibit highly concentrated and distinct personality trait distributions. For instance, the purple cluster is characterized by high Agreeableness (Agr), while the orange cluster displays notably high levels of Extraversion (Ext) and Neuroticism (Neu). These observations underscore the effectiveness of PRAC, through PreferSelect and PreferMerge, in modeling personalized aesthetic preferences.

\begin{figure}[t]
	\centering
	\includegraphics[width=0.97\columnwidth]{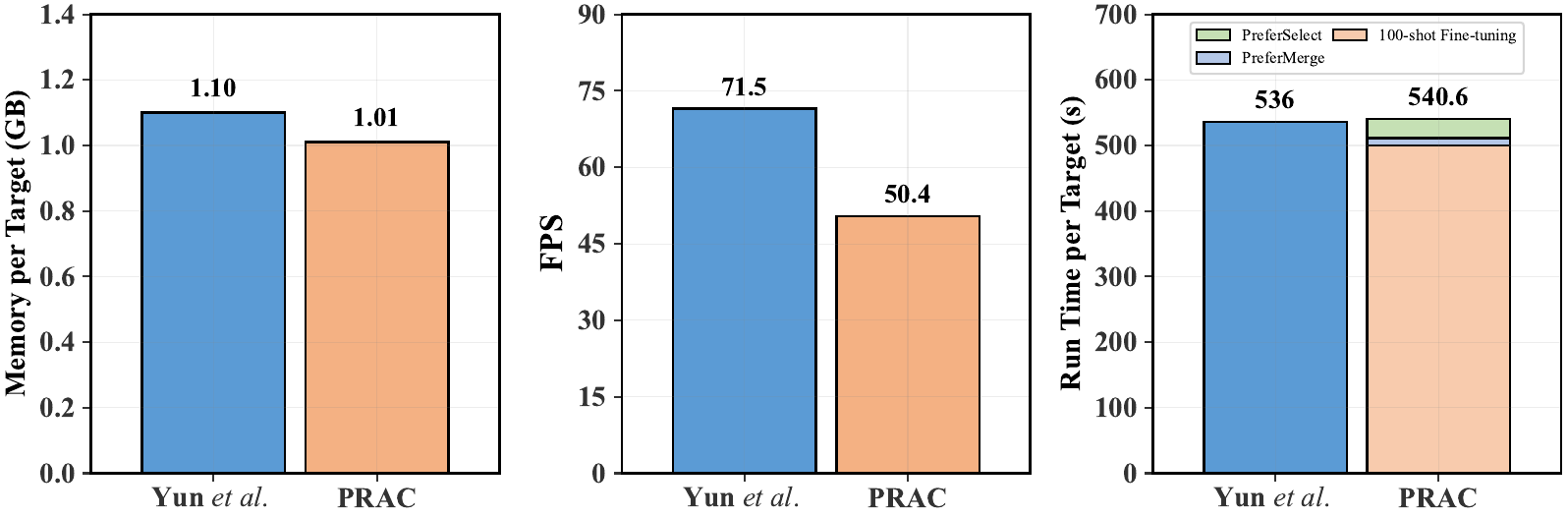} 
	\caption{Comparison of computational costs between PRAC and the current state-of-the-art PIAA method \protect\cite{yun2024scaling}.}
	\label{fig7}
\end{figure}

\subsection{Computational Cost}
\label{sec:cost}
For the proposed PRAC, all target users share a frozen MLLM backbone (e.g., mPLUG-Owl3) equipped with different trained personalized LoRAs. Figure \ref{fig7} compares the computational costs (memory per target, training time per target, and inference speed) between PRAC and the current state-of-the-art ViT-based PIAA model \cite{yun2024scaling}. The results demonstrate that PRAC achieves competitive performance in both memory consumption and runtime efficiency. All metrics are measured on a single RTX 3090 GPU.

\section{Conclusion}
In this paper, we rethink personalized image aesthetic assessment by modeling preference richness across images and cross-user preference similarity. We have proposed PRAC, the first MLLM-based PIAA model that accurately predicts users' personalized aesthetic ratings on images and provides interpretable rationales explaining divergent judgments. Through preference-rich sample mining and aesthetically-resonant cohort merging, we effectively model individual aesthetic preferences under few-shot annotations. Extensive evaluations across four PIAA benchmark datasets demonstrate the superior performance of the proposed model. While very encouraging results have been achieved in this work, investigations on the cohort-relatedness of aesthetic preferences would further facilitate preference modeling, which is worth of more explorations.

\begin{acks}
This work is supported by National Natural ScienceFoundation of China under Grants 62471349, 62171340,62301378, and 625B2142, Fundamental Research Fundsfor the Central Universities under Grant QTZX25076and YJSJ25004, the China Postdoctoral Science Foundation under Grant 2024M762553.
\end{acks}

\bibliographystyle{ACM-Reference-Format}
\bibliography{acmmm26}

\end{document}